\documentclass[nohyperref]{article}

\usepackage[accepted]{icml2022}
\usepackage{microtype}
\usepackage{graphicx}
\usepackage{subfigure}
\usepackage{longtable}
\usepackage{booktabs} 
\usepackage{float}

\usepackage{hyperref}

\usepackage{tikz}

\usepackage{amsmath}
\usepackage{amssymb}
\usepackage{mathtools}
\usepackage{amsthm}
\usepackage{enumitem} 

\usepackage[capitalize,noabbrev]{cleveref}

\usepackage{mathtools}

\theoremstyle{plain}
\newtheorem{theorem}{Theorem}[section]

\theoremstyle{definition}
\newtheorem{definition}[theorem]{Definition}

\theoremstyle{remark}

\usepackage[textsize=tiny]{todonotes}

\usepackage{alexdefs}
\usepackage[subtle]{savetrees}

\newcommand{\trueg}{{G^\star}}
\newcommand{\unor}{\mathcal{U}(\mathcal{M}_\trueg)}
\newcommand{\Mexpert}[1]{\Mcal^{E}_{#1}}
\newcommand{\MexpertS}[1]{\Mcal^{E,S}_{#1}}

\icmltitlerunning{Causal Discovery with Language Models as Imperfect Experts}

\begin{document}

\twocolumn[
\icmltitle{Causal Discovery with Language Models as Imperfect Experts}



\icmlsetsymbol{equal}{*}

\begin{icmlauthorlist}
\icmlauthor{Stephanie Long}{equal,mcgill}
\icmlauthor{Alexandre Pich\'e}{equal,snow,udem,mila}
\icmlauthor{Valentina Zantedeschi}{equal,snow}
\icmlauthor{Tibor Schuster}{mcgill}
\icmlauthor{Alexandre Drouin}{snow,mila}
\end{icmlauthorlist}

\icmlaffiliation{mila}{Mila - Quebec AI Institute}
\icmlaffiliation{udem}{Universit\'{e} de Montr\'{e}al}
\icmlaffiliation{snow}{ServiceNow Research}
\icmlaffiliation{mcgill}{McGill University}

\icmlcorrespondingauthor{Valentina Zantedeschi}{vzantedeschi@gmail.com}

\icmlkeywords{Machine Learning, ICML}

\vskip 0.3in
]



\printAffiliationsAndNotice{\icmlEqualContribution} 

\begin{abstract}
Understanding the causal relationships that underlie a system is a fundamental prerequisite to accurate decision-making. In this work, we explore how expert knowledge can be used to improve the data-driven identification of causal graphs, beyond Markov equivalence classes. In doing so, we consider a setting where we can query an expert about the orientation of causal relationships between variables, but where the expert may provide erroneous information. We propose strategies for amending such expert knowledge based on consistency properties, e.g., acyclicity and conditional independencies in the equivalence class. We then report a case study, on real data, where a large language model is used as an imperfect expert.
\end{abstract}

\section{Introduction}

Understanding the cause-and-effect relationships that underlie a complex system is critical to accurate decision-making.
Unlike any statistical association, causal relationships allow us to anticipate the system's response to interventions. 
Currently, randomized control trials (RCTs) serve as the gold standard for establishing causation~\citep{peters2017elements}. 
However, RCTs can be costly and oftentimes impractical or unethical.
As such, there has been growing interest in \emph{causal discovery}, which aims to uncover causal relationships from data collected by passively observing a system (see \citep{cd_review} for a review).

Causal discovery methods have been successfully applied in various fields, including genetics~\citep{sachs2005causal} and climate science~\citep{runge2019inferring}. 
Nevertheless, a fundamental limitation of such methods is their ability to only recover the true graph of causal relationships up to a set of equivalent solutions known as the \emph{Markov equivalence class} (MEC),
leading to uncertainty in downstream applications, such as estimating the effect of interventions~\citep{maathuis2009}.

One approach to reducing such uncertainty is the incorporation of expert knowledge, e.g., to rule out the existence of certain edges and reduce the set of possible solutions~\citep{meek1995}.
However, such methods typically assume that the knowledge provided by the \emph{expert is correct}.
In this work, we consider a more realistic case, where the \emph{expert is potentially incorrect}.
Our approach leverages such \emph{imperfect experts}, e.g., large language models, to reduce uncertainty in the output of a causal discovery algorithm by orienting edges, while maintaining fundamental consistency properties, such as the acyclicity of the causal graph and the conditional independencies in the MEC.

\textbf{Contributions:}
\vspace{-3mm}
\begin{itemize}
\setlength\itemsep{0pt}
    \item We formalize the use of \emph{imperfect experts} in causal discovery as an optimization problem that minimizes the size of the MEC while ensuring that the true graph is still included (\cref{sec:problem}).
    \item We propose a greedy approach that relies on Bayesian inference to optimize this objective by incrementally incorporating expert knowledge (\cref{sec:methods}).
    \item We empirically evaluate the performance of our approach, on real data, with an expert that returns correct orientations with some fixed probability (\cref{sec:experiments}).
    \item We then empirically assess if the approach holds when taking a large language model as the expert -- with mitigated results (\cref{sec:experiments}).
\end{itemize}




\section{Background and  Related Work}\label{sec:background}

We now review key background concepts and related work.

\textbf{Causal Bayesian networks:~~}
Let $\Xb \coloneqq (X_1, \dots, X_d)$ be a vector of $d$ random variables with distribution $p(\Xb)$ and $\trueg \coloneqq \left<\Vcal_\trueg, \Ecal_\trueg \right>$ be a directed acyclic graph (DAG) with vertices ${\Vcal_\trueg = \{v_1, \dots, v_d\}}$ and edges $\Ecal_\trueg \subset \Vcal_\trueg \times \Vcal_\trueg$. Each vertex $v_i \in \Vcal_\trueg$ corresponds to a random variable $X_i$ and a directed edge $(v_i, v_j) \in \Ecal_\trueg$ represents a direct causal relationship from $X_i$ to $X_j$.
We assume that $p(\Xb)$ is \emph{Markovian} with respect to $\trueg$, i.e.,
\vspace{-1mm}
$$p(X_1, \ldots, X_d) = \prod_{i=1}^d p(X_i \mid \text{pa}_i^\trueg),$$
\vspace{-1mm}
where $\text{pa}_i^\trueg$ denotes the parents of $X_i$ in $\trueg$.

\textbf{Causal discovery:~} This task consists of recovering $\trueg$ from data, which are typically sampled from $p(\Xb)$~\citep{cd_review}.
Existing methods can be broadly classified as being: i) constraint-based~\citep{spirtes2000constructing,FCI}, which use conditional independence tests to rule-out edges, or ii) score-based~\citep{chickering2002optimal,zheng2018dags}, which search for the DAG that optimizes some scoring function.
One common limitation of these approaches is their inability to fully identify the true underlying graph $\trueg$ beyond its \emph{Markov equivalence class} (MEC)~\citep{peters2017elements}.

\textbf{Equivalence classes:~~} 
The MEC, $\Mcal_\trueg$, is a set of graphs that includes $\trueg$ and all other DAGs with equivalent conditional independences. These may have different edge orientations, leading to uncertainty in downstream tasks, such as treatment effect estimation \citep{maathuis2009}. 
One common approach to reducing the size of $\Mcal_\trueg$ is to include interventional data ~\citep{eberhardt2005,brouillard2020differentiable,mooij2020joint}.
However, similar to RCTs, the collection of such data may not always be feasible or ethical. 
An alternative approach, which we adopt in this work, is to eliminate graphs that are deemed implausible by an expert.


\textbf{Expert knowledge:~~} 
Previous work has considered experts that give: (i) forbidden edges~\citep{meek1995}, (ii) (partial) orderings of the variables~\citep{scheines1998tetrad,andrews2020}, (iii) ancestral constraints~\citep{decampos2007bayesian,li2018bayesian,chen2016learning}, and (iv) constraints on interactions between types of variables~\citep{brouillard2022typing} 
(see \citet{constantinou2023impact} for a review).
Typically, all DAGs that are contradicted by the expert are discarded, resulting in a new equivalence class $\Mexpert{~} \subseteq \Mcal_\trueg$.
One pitfall is that, realistically, an expert is unlikely to always be correct, and thus, $\trueg$ might be discarded, i.e., $\trueg \not\in \Mexpert{~}$.
In this work, we attempt to reduce $\Mcal_\trueg$ as much as possible, while ensuring $\trueg \in \Mexpert{~}$ with high probability, in the presence of \emph{imperfect expert knowledge}.
We note that our work is akin to \citet{oates2017}, but a key difference is that they assume a \emph{directionally informed} expert, i.e., that cannot misorient edges in $\trueg$. 
Moreover, their approach is \emph{expert-first}, i.e., data is used to expand an initial graph given by an expert,
while our approach is \emph{data-first}, i.e., the expert is used to refine the solution of a causal discovery algorithm.

\textbf{Large language models:~~}
In situations where access to human experts is limited, Large Language Models~(LLMs), such as GPT-4~\citep{openai2023gpt4}, offer promising alternatives.
Recent studies have demonstrated that certain LLMs possess a rich knowledge base that encompasses valuable information for causal discovery~\citep{lmpriors, long2023, hobbhahn2022investigating, willig2022can, kiciman2023causal, tu2023causaldiscovery}, achieving state-of-the-art accuracy on datasets such as the Tübingen pairs~\citep{JMLR:v17:14-518}.
In this work, we investigate the use of LLMs as \emph{imperfect experts} within the context of causal discovery.
Unlike prior approaches, which typically assume the correctness of extracted knowledge,\footnote{The Bayesian approach of \citet{lmpriors} is an exception.} we propose strategies to use, potentially incorrect, LLM knowledge to eliminate some graphs in $\Mcal_\trueg$, while ensuring that $\trueg \in \Mexpert{~}$ with high probability.

\section{Problem Setting}\label{sec:problem}

We now formalize our problem of interest.
Let $\trueg$ represent the true causal DAG, as defined in \cref{sec:background}, and let $\Mcal_\trueg$ be its MEC. 
We assume that $\Mcal_\trueg$ is known, e.g., that it has been obtained via some causal discovery algorithm.
Further, we assume the availability of \emph{metadata} $\{\mu_1, \ldots, \mu_d\}$, where each $\mu_i$ provides some information about $X_i$, e.g., a name, a brief description, etc.
We then assume access to an expert, who consumes such metadata and makes decisions:
\begin{definition}
\textbf{(Expert)} An expert is a function that, when queried with the metadata for a pair of variables $(\mu_i, \mu_j)$, returns a hypothetical orientation for the $X_i - X_j$ edge:
\vspace{-1mm}
\begin{equation}
    E\left(\mu_i, \mu_j \right) = \left\{
    \begin{array}{ll}
    		\rightarrow  & \mbox{if it believes that } (v_i, v_j) \in \Ecal_\trueg \\
    		\leftarrow & \mbox{if it believes that } (v_j, v_i) \in \Ecal_\trueg\\
    \end{array}\right..
\end{equation}
\end{definition}
\vspace{-2mm}

Of note, $E\left(\mu_i, \mu_j \right)$ can be incorrect (\textit{imperfect expert}) and thus, our problem of interest consists in
elaborating strategies to maximally make use of such imperfect knowledge.


Let $\unor$ be the set of indices of all pairs of variables related by an edge whose orientation is ambiguous in $\Mcal_\trueg$:
\begin{equation}
    \begin{split}
        \unor \,\coloneqq\, \left\{ (i, j) \mid i < j \text{ and } \exists G, G' \in \Mcal_\trueg \text{ s.t. }\right.\\
        \left. (v_i, v_j) \in \Ecal_{G} \,\wedge\, (v_j, v_i) \in \Ecal_{G'} \right\}.\\
    \end{split}
\end{equation}
We aim to elaborate a strategy $S$ that uses the expert's knowledge to orient edges in $\unor$ and obtain a new equivalence class $\MexpertS{~}$, such that uncertainty is reduced to the minimum, i.e., $\lvert \MexpertS{~} \lvert \ll \lvert\Mcal_\trueg\lvert$, but $\trueg$ still belongs to $\MexpertS{~}$ with high probability, that is:
\begin{align}\label{eq:problem-full-graph}
    \min \ & \ \left\lvert\MexpertS{~}\right\lvert\\
    \text{such that} & \ p\left(\trueg \in \MexpertS{~} \right) \geq 1 - \eta \nonumber,
\end{align}

where $\eta \in [0, 1]$ quantifies tolerance to the risk that the true graph $\trueg$ is not in the resulting equivalence class.
This problem can be viewed as a trade-off between reducing uncertainty, by shrinking the set of plausible DAGs, and the risk associated with making decisions based on an imperfect expert.

\section{Strategies for Imperfect Experts}\label{sec:methods}

Instead of blindly accepting expert orientations, we leverage the consistency information provided by the true MEC to estimate which decisions are most likely incorrect. 
Indeed, among all possible combinations of edge orientations, only a few are possible, since many of them would create cycles or introduce new v-structures.
The different strategies that we now propose for solving Problem~\eqref{eq:problem-full-graph} leverage such consistency imperatives, as well as Bayesian inference, to increase robustness to errors in expert knowledge.

\textbf{Noise model:~~}
First, let us define the noise model that, we assume, characterizes mistakes made by the expert.
Figure~\ref{fig:dep-graph} shows the dependency graph for the decision process of a type of imperfect expert that we dub ``$\varepsilon$-expert".
For any pair $p_i = (p_{i1},\; p_{i2}) \in \unor$, we use the notation $O_{p_i}$ to denote the \emph{unknown} true edge orientation and $E_{p_i} \coloneqq E(\mu_{p_{i1}}, \mu_{p_{i2}})$ denotes the orientation given by the expert.
Further, for any subset of indices $I \subseteq \unor$, we use $O_I \coloneqq \{ O_{p_i} \}_{i=1}^{|I|}$; the same applies to $E_I$.
Notice that (i) true edge orientations are, in general, interdependent because of the aforementioned consistency properties of the MEC, and that (ii) edges already oriented in $\Mcal_\trueg$ are not represented since they are constants (i.e., the expert is not queried for those).
In this model, we assume that, for any $p_i \in \unor$, the expert's response depends only on the true value $O_{p_i}$, i.e., $p(E_{p_i} \mid O_{\unor}) = p(E_{p_i} \mid O_{p_i})$ and is incorrect with constant probability $\varepsilon$.

We now define the components of our Bayesian approach.

\textbf{Prior:~~}We consider a simple prior that encodes the knowledge given by the true MEC.
It boils down to an uniform prior over the graphs in $\Mcal_\trueg$, effectively assigning no mass to any edge combination that is not consistent (creates a cycle or a new v-structure). 
Thus, the prior for a partial edge orientation $O_I$ corresponds to its frequency in the graphs of $\Mcal_\trueg$:
$$p(O_I) \mspace{5mu} = \mspace{5mu} \sum_{\ob_{\neg I}} \mspace{5mu} p\left(O_I, \mspace{5mu} O_{\unor \setminus I} \mspace{5mu} = \mspace{5mu} \ob_{\neg I} \right),$$
where we marginalize over all possible combinations of values, $\ob_{\neg I}$, for the remaining unoriented edges $O_{\unor \setminus I}$.

\textbf{Posterior:~} The posterior probability that orientations for all edges in $\unor$ are correct, given all observed expert decisions $E_{\unor}$, is then given by:
\begin{equation}\label{eq:p_g_in_mec}
    \begin{split} 
        p\left(O_{\unor} \mid E_{\unor}\right) = \\ 
        & \mspace{-100mu} \frac{p\left(E_{\unor} \mid O_{\unor}\right) \mspace{5mu} p\left(O_{\unor}\right)}{p\left(E_{\unor}\right)}
    \end{split},
\end{equation}
where, for the $\varepsilon$-expert noise model, the likelihood is s.t.,
$$p(E_{\unor} \mid O_{\unor}) \quad = \prod_{p_i \in \unor} p(E_{p_i} \mid O_{p_i}).$$
In contrast, due to interdependencies between the true edge orientations, the posterior probability cannot similarly be factorized and, in general, $p(O_{p_i} \mid E_{\unor}) \neq p(O_{p_i} \mid E_{p_i})$.
Note that the posterior for a subset edges $I \subseteq \unor$, e.g, oriented by an iterative strategy, can be obtained via simple marginalization.
Finally, note that the posterior can be used to estimate $p\left(\trueg \in \MexpertS{~}\right)$, since any mistake in orienting $p_i \in \unor$ results in excluding $\trueg$ from $\MexpertS{~}$.

\usetikzlibrary{graphs}

\begin{figure}[htp]
    \centering
    \vspace*{15mm}
    \resizebox{0.23\textwidth}{!}{
    \begin{tikzpicture}
        \tikz \graph [math nodes] {
            O_{p_1} -- O_{p_2} -- "\dots" -- O_{p_u},
            { [fresh nodes] E_{p_1} -!- E_{p_2} -!- "\dots" -!- E_{p_u}},
            O_{p_1} --[bend left] O_{p_u},
            O_{p_2} --[bend left] O_{p_u},
            O_{p_1} -> E_{p_1};
            O_{p_2} -> E_{p_2};
            O_{p_u} -> E_{p_u};
        };
    \end{tikzpicture}}
    \vspace*{-15mm}
    \caption{The $\epsilon$-expert's dependency graph between true edge orientations ($O_{p_i}$) and expert decisions ($E_{p_i}$), where $u = |\unor|$.}
    \label{fig:dep-graph}
\end{figure}
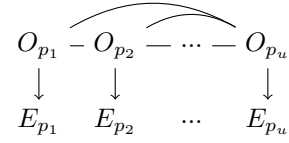

\textbf{Greedy approach:~~}
We now propose a greedy strategy for optimizing Problem~\eqref{eq:problem-full-graph} that iteratively orients edges in $\unor$.
Let $\Mcal^{(t)}$ denote the MEC at the $t$-th iteration of the algorithm and let $\Mcal^{(t)}_{p_i}$ denote the MEC resulting from additionally orienting $p_i \in \Ucal(\Mcal^{(t)})$ at step $t$ and propagating any consequential orientations using \citet{meek1995}'s rules.
The algorithm starts with $\Mcal^{(1)} = \Mcal_\trueg$.
We consider two strategies to greedily select the best $p_i$:
\vspace{-2mm}
\begin{enumerate}
    \setlength\itemsep{-6pt}
    \item $S_\text{size}$: selects the edge that leads to the smallest equivalence class:\vspace{-3mm}
    $$\arg\min_{p_i} ~~ \lvert \Mcal^{(t)}_{p_i} \lvert $$
    \item $S_\text{risk}$: selects the edge that leads to the lowest risk of excluding $\trueg$ from the equivalence class:
    $$\arg\min_{p_i} \left[ 1 - p\left(O_{\unor \setminus \Ucal\left(\Mcal^{(t)}_{p_i}\right)} \mid E_{\unor}\right) \right]$$
\end{enumerate}
\vspace{-1mm}
This procedure is repeated while $p\left(\trueg \in \Mcal^{(t)}_{p_i}\right)$, estimated according to \cref{eq:p_g_in_mec}, is greater or equal to $1 - \eta$.
\begin{figure*}[t]
    \centering
    \includegraphics[width=\textwidth]{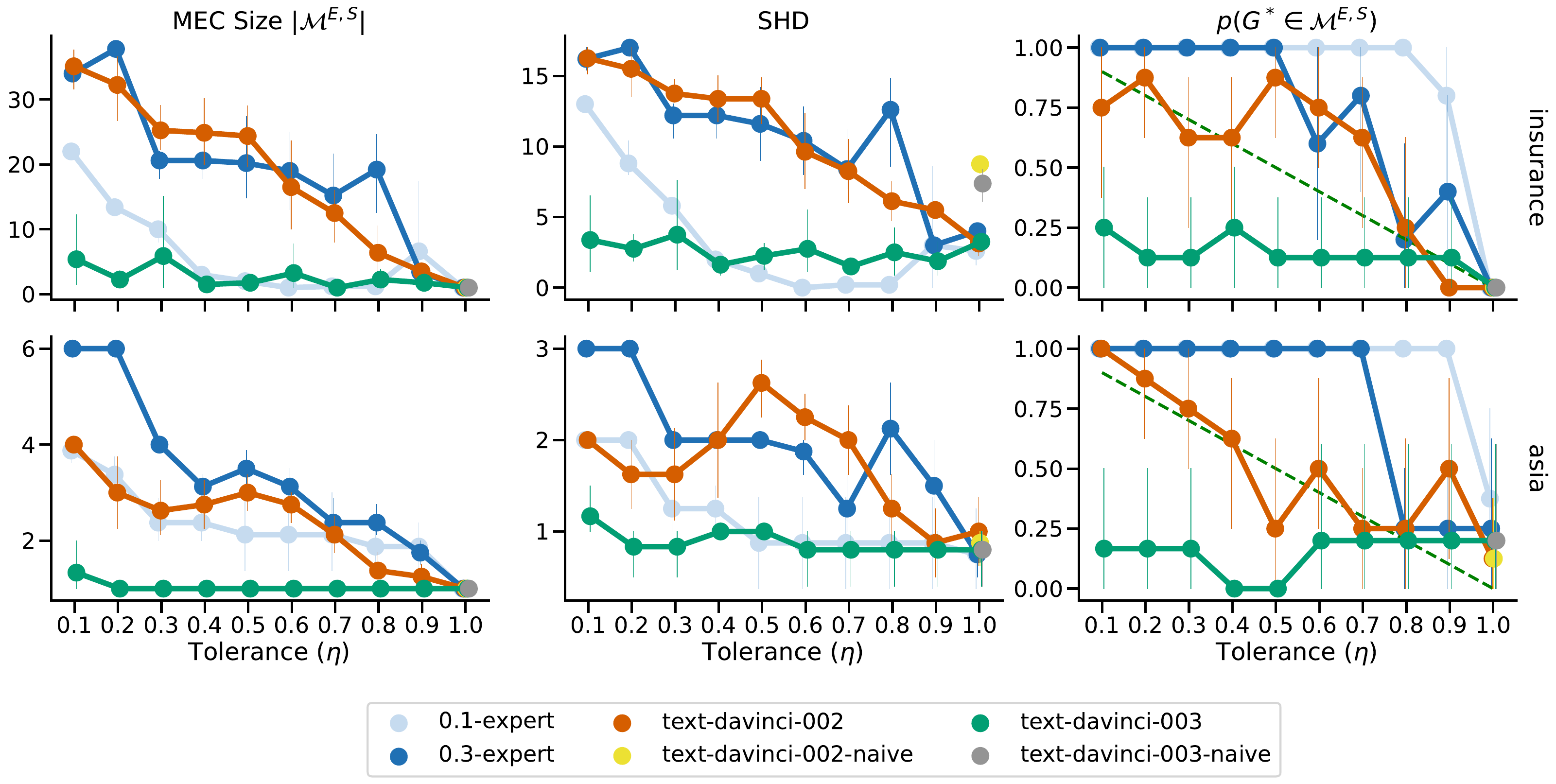}
    \vspace{-2mm}
    \caption{Results for \texttt{Insurance} and \texttt{Asia} and strategy $S_\text{risk}$. For the $\varepsilon$-experts, we consider $\varepsilon \in \{0.1, 0.3\}$. For the LLM-based experts, we consider the \texttt{text-davinci-}$\{$\texttt{002}, \texttt{003}$\}$ versions of GPT-3.5. We also report results for \emph{naive} variants that follow \citet{long2023} and do not make use of our greedy approach. Error bars show a $95\%$ confidence interval. For all experts, except for \texttt{text-davinci-003}, both the MEC size and SHD decrease as tolerance is increased and $p(G^\star\in \mathcal{M}^{E,S}) \geq 1-\eta$.
    }
    \label{fig:main-text-results}
\end{figure*}


\section{Results and Discussion}\label{sec:experiments}
We now evaluate the ability of our approach to leverage imperfect expert knowledge using real-world causal Bayesian networks from the \emph{bnlearn} repository~\citep{bnlearn}.

\textbf{Networks:~~}We considered the following networks:
(i) \texttt{Asia}~\citep{asia}, (ii) \texttt{ALARM}~\citep{alarm}, (iii) \texttt{CHILD}~\citep{child}, and (iv) \texttt{Insurance}~\citep{insurance}.
For each network, we extracted variable descriptions from the related publication and used them as metadata $\mu_i$ (see \cref{app:bnlearn}).

\textbf{Experts:~~} We considered two kinds of expert: (i) $\varepsilon$-experts, as defined in \cref{sec:methods}, with various levels $\varepsilon$, and (ii) LLM-based experts based on GPT-3.5~\citep{ouyang2022training}. Details about prompting can be found at \cref{app:llm-experts}. For each kind of expert, we considered both strategies: $S_\text{size}$ and $S_\text{risk}$. Moreover, for the LLM-based expert, we also considered a naive strategy that consists of simply orienting all edges according to the expert, as in \citet{long2023}.

\textbf{Metrics:~~}
The expert/strategy combinations were evaluated based on: (i) the resulting size of their equivalence class, $\lvert \MexpertS{~} \lvert$, (ii) the structural Hamming distance (SHD) between the completed partially DAG (CP-DAG; see \citet{cd_review}) of $\MexpertS{~}$ and the true graph $\trueg$, (iii) an empirical estimate of $p(\trueg \in \MexpertS{~})$, taken over repetitions of the experiment.

\textbf{Protocol:} For each Bayesian network, we extracted the MEC $\Mcal_\trueg$ based on the structure of $\trueg$. This simulates starting from the ideal output of a causal discovery algorithm. We then attempted to reduce the size of $\Mcal_\trueg$ by querying each expert according to the greedy approach in \cref{sec:methods} for $\eta \in [0.1, 1]$, where $\eta = 1$ corresponds to disregarding the constraint of Problem~\ref{eq:p_g_in_mec}. Each $\varepsilon-$expert experiment was repeated $5$ times and LLM-expert experiment was repeated 8 times. Given that the LLM-experts have a deterministic output for a given prompt, we randomized the causation verb in order to introduce stochasticity (see \cref{app:llm-experts}). \cref{fig:main-text-results} shows the results of our experiments for \texttt{Insurance} and \texttt{Asia} with strategy $S_\text{risk}$. Results for other networks and strategy $S_\text{size}$ are in \cref{app:more-results}.

\textbf{Results for $\varepsilon$-experts:~}
On all networks, our approach, combined with both strategies, decreases the MEC's size for all noise levels ($\varepsilon$), while keeping the true graph in $\MexpertS{~}$ with probability at least $1-\eta$, as predicted by our theoretical results. Consequently, the SHD also decreases as the tolerance to risk increases. This highlights the effectiveness of our approach when the expert satisfies the noise model of \cref{sec:methods}.


\textbf{Results for the LLM-expert:}
Overall, we observe a clear reduction in SHD for $\MexpertS{~}$ compared to the starting point $\Mcal_\trueg$.
This shows that some causally-relevant knowledge can be extracted from LLMs, which is in line with the conclusions of recent work.

On all datasets, the LLM-based experts achieve SHDs that are on par or better than those of their naive counterparts~\citep{long2023} for $\eta = 1$, 
while additionally enabling the control of the probability of excluding $\trueg$.
Further, on every dataset, except for \texttt{ALARM}, each LLM-based expert performs comparably to at least one of the $\varepsilon$-experts.
For \texttt{ALARM}, we observe that $\trueg$ is excluded from $\MexpertS{}$, even for small tolerance $\eta$.
This can be explained by ambiguities in the metadata, which are sometimes ambiguous even for human experts (see \cref{app:bnlearn}).

Finally, another key observation is the poor uncertainty calibration of \texttt{text-davinci-003} compared to \texttt{text-davinci-002}, which is in line with observations made by \citet{openai2023gpt}.
The \texttt{text-davinci-003} model is often over-confident in its answers, which leads it to underestimate the probability of excluding $\trueg$ from $\MexpertS{}$.
Consequently, even for small tolerance $\eta$, the resulting equivalence classes contain incorrectly oriented edges.

\section{Conclusion}

This work studied how imperfect expert knowledge can be used to refine the output of causal discovery algorithms.
We proposed a greedy algorithm that iteratively rejects graphs from a MEC, while controlling the probability of excluding the true graph.
Our empirical study revealed that our approach is effective when combined with experts that satisfy our assumptions.
However, its performance was mitigated when a LLM was used as the expert.
Nevertheless, our results show the clear potential of LLMs to aid causal discovery and we believe that further research in this direction is warranted.
Possible extensions to this work include the exploration of noise models better-suited for LLMs, as well as alternative methods for querying such models (e.g., different prompt styles, better uncertainty calibration, etc.,).
Further, our approach could be coupled with Bayesian causal discovery methods, replacing our MEC-based prior by one derived from a learned posterior distribution over graphs.


\bibliography{references}

\begin{thebibliography}{38}
\providecommand{\natexlab}[1]{#1}
\providecommand{\url}[1]{\texttt{#1}}
\expandafter\ifx\csname urlstyle\endcsname\relax
  \providecommand{\doi}[1]{doi: #1}\else
  \providecommand{\doi}{doi: \begingroup \urlstyle{rm}\Url}\fi

\bibitem[Andrews et~al.(2020)Andrews, Spirtes, and Cooper]{andrews2020}
Andrews, B., Spirtes, P., and Cooper, G.~F.
\newblock On the completeness of causal discovery in the presence of latent
  confounding with tiered background knowledge.
\newblock In Chiappa, S. and Calandra, R. (eds.), \emph{Proceedings of the
  Twenty Third International Conference on Artificial Intelligence and
  Statistics}, volume 108 of \emph{Proceedings of Machine Learning Research},
  pp.\  4002--4011. PMLR, 26--28 Aug 2020.
\newblock URL \url{https://proceedings.mlr.press/v108/andrews20a.html}.

\bibitem[Bai et~al.(2022)Bai, Kadavath, Kundu, Askell, Kernion, Jones, Chen,
  Goldie, Mirhoseini, McKinnon, et~al.]{bai2022constitutional}
Bai, Y., Kadavath, S., Kundu, S., Askell, A., Kernion, J., Jones, A., Chen, A.,
  Goldie, A., Mirhoseini, A., McKinnon, C., et~al.
\newblock Constitutional ai: Harmlessness from ai feedback.
\newblock \emph{arXiv preprint arXiv:2212.08073}, 2022.

\bibitem[Beinlich et~al.(1989)Beinlich, Suermondt, Chavez, and Cooper]{alarm}
Beinlich, I.~A., Suermondt, H.~J., Chavez, R.~M., and Cooper, G.~F.
\newblock The alarm monitoring system: A case study with two probabilistic
  inference techniques for belief networks.
\newblock pp.\  247--256, 1989.

\bibitem[Binder et~al.(1997)Binder, Koller, Russell, and Kanazawa]{insurance}
Binder, J., Koller, D., Russell, S., and Kanazawa, K.
\newblock Adaptive probabilistic networks with hidden variables.
\newblock \emph{Machine Learning}, 29\penalty0 (2-3):\penalty0 213--244, 1997.

\bibitem[Brouillard et~al.(2020)Brouillard, Lachapelle, Lacoste,
  Lacoste-Julien, and Drouin]{brouillard2020differentiable}
Brouillard, P., Lachapelle, S., Lacoste, A., Lacoste-Julien, S., and Drouin, A.
\newblock Differentiable causal discovery from interventional data.
\newblock \emph{Advances in Neural Information Processing Systems},
  33:\penalty0 21865--21877, 2020.

\bibitem[Brouillard et~al.(2022)Brouillard, Taslakian, Lacoste, Lachapelle, and
  Drouin]{brouillard2022typing}
Brouillard, P., Taslakian, P., Lacoste, A., Lachapelle, S., and Drouin, A.
\newblock Typing assumptions improve identification in causal discovery.
\newblock In \emph{Conference on Causal Learning and Reasoning}, pp.\
  162--177. PMLR, 2022.

\bibitem[Chen et~al.(2016)Chen, Shen, Choi, and Darwiche]{chen2016learning}
Chen, E. Y.-J., Shen, Y., Choi, A., and Darwiche, A.
\newblock Learning bayesian networks with ancestral constraints.
\newblock \emph{Advances in Neural Information Processing Systems}, 29, 2016.

\bibitem[Chickering(2002)]{chickering2002optimal}
Chickering, D.~M.
\newblock Optimal structure identification with greedy search.
\newblock \emph{Journal of machine learning research}, 3\penalty0
  (Nov):\penalty0 507--554, 2002.

\bibitem[Choi et~al.(2022)Choi, Cundy, Srivastava, and Ermon]{lmpriors}
Choi, K., Cundy, C., Srivastava, S., and Ermon, S.
\newblock Lmpriors: Pre-trained language models as task-specific priors.
\newblock \emph{arXiv preprint arXiv: 2210.12530}, 2022.

\bibitem[Constantinou et~al.(2023)Constantinou, Guo, and
  Kitson]{constantinou2023impact}
Constantinou, A.~C., Guo, Z., and Kitson, N.~K.
\newblock The impact of prior knowledge on causal structure learning.
\newblock \emph{Knowledge and Information Systems}, pp.\  1--50, 2023.

\bibitem[de~Campos \& Castellano(2007)de~Campos and
  Castellano]{decampos2007bayesian}
de~Campos, L.~M. and Castellano, J.~G.
\newblock Bayesian network learning algorithms using structural restrictions.
\newblock \emph{International Journal of Approximate Reasoning}, 45\penalty0
  (2):\penalty0 233--254, 2007.

\bibitem[Eberhardt et~al.(2005)Eberhardt, Glymour, and Scheines]{eberhardt2005}
Eberhardt, F., Glymour, C., and Scheines, R.
\newblock On the number of experiments sufficient and in the worst case
  necessary to identify all causal relations among n variables.
\newblock In \emph{Conference on Uncertainty in Artificial Intelligence}, 2005.

\bibitem[Glymour et~al.(2019)Glymour, Zhang, and Spirtes]{cd_review}
Glymour, C., Zhang, K., and Spirtes, P.
\newblock Review of causal discovery methods based on graphical models.
\newblock \emph{Frontiers in Genetics}, 10, 2019.
\newblock ISSN 1664-8021.
\newblock \doi{10.3389/fgene.2019.00524}.
\newblock URL
  \url{https://www.frontiersin.org/articles/10.3389/fgene.2019.00524}.

\bibitem[Hobbhahn et~al.(2022)Hobbhahn, Lieberum, and
  Seiler]{hobbhahn2022investigating}
Hobbhahn, M., Lieberum, T., and Seiler, D.
\newblock Investigating causal understanding in llms.
\newblock 2022.

\bibitem[Kadavath et~al.(2022)Kadavath, Conerly, Askell, Henighan, Drain,
  Perez, Schiefer, Dodds, DasSarma, Tran-Johnson, et~al.]{kadavath2022language}
Kadavath, S., Conerly, T., Askell, A., Henighan, T., Drain, D., Perez, E.,
  Schiefer, N., Dodds, Z.~H., DasSarma, N., Tran-Johnson, E., et~al.
\newblock Language models (mostly) know what they know.
\newblock \emph{arXiv preprint arXiv:2207.05221}, 2022.

\bibitem[K{\i}c{\i}man et~al.(2023)K{\i}c{\i}man, Ness, Sharma, and
  Tan]{kiciman2023causal}
K{\i}c{\i}man, E., Ness, R., Sharma, A., and Tan, C.
\newblock Causal reasoning and large language models: Opening a new frontier
  for causality.
\newblock \emph{arXiv preprint arXiv:2305.00050}, 2023.

\bibitem[Lauritzen \& Spiegelhalter(1988)Lauritzen and Spiegelhalter]{asia}
Lauritzen, S.~L. and Spiegelhalter, D.~J.
\newblock Local computation with probabilities on graphical structures and
  their application to expert systems (with discussion).
\newblock \emph{Journal of the Royal Statistical Society: Series B (Statistical
  Methodology)}, 50\penalty0 (2):\penalty0 157--224, 1988.

\bibitem[Li \& Beek(2018)Li and Beek]{li2018bayesian}
Li, A. and Beek, P.
\newblock Bayesian network structure learning with side constraints.
\newblock In \emph{International conference on probabilistic graphical models},
  pp.\  225--236. PMLR, 2018.

\bibitem[Long et~al.(2023)Long, Schuster, and Piché]{long2023}
Long, S., Schuster, T., and Piché, A.
\newblock Can large language models build causal graphs?
\newblock \emph{arXiv preprint arXiv: 2303.05279}, 2023.

\bibitem[Maathuis et~al.(2009)Maathuis, Kalisch, and
  B{\"u}hlmann]{maathuis2009}
Maathuis, M.~H., Kalisch, M., and B{\"u}hlmann, P.
\newblock {Estimating high-dimensional intervention effects from observational
  data}.
\newblock \emph{The Annals of Statistics}, 37\penalty0 (6A):\penalty0 3133 --
  3164, 2009.
\newblock \doi{10.1214/09-AOS685}.
\newblock URL \url{https://doi.org/10.1214/09-AOS685}.

\bibitem[Meek(1995)]{meek1995}
Meek, C.
\newblock Causal inference and causal explanation with background knowledge.
\newblock In \emph{Proceedings of the Eleventh Conference on Uncertainty in
  Artificial Intelligence}, UAI'95, pp.\  403–410, San Francisco, CA, USA,
  1995. Morgan Kaufmann Publishers Inc.
\newblock ISBN 1558603859.

\bibitem[Mooij et~al.(2016)Mooij, Peters, Janzing, Zscheischler, and
  Sch{{\"o}}lkopf]{JMLR:v17:14-518}
Mooij, J.~M., Peters, J., Janzing, D., Zscheischler, J., and Sch{{\"o}}lkopf,
  B.
\newblock Distinguishing cause from effect using observational data: Methods
  and benchmarks.
\newblock \emph{Journal of Machine Learning Research}, 17\penalty0
  (32):\penalty0 1--102, 2016.
\newblock URL \url{http://jmlr.org/papers/v17/14-518.html}.

\bibitem[Mooij et~al.(2020)Mooij, Magliacane, and Claassen]{mooij2020joint}
Mooij, J.~M., Magliacane, S., and Claassen, T.
\newblock Joint causal inference from multiple contexts.
\newblock \emph{The Journal of Machine Learning Research}, 21\penalty0
  (1):\penalty0 3919--4026, 2020.

\bibitem[Oates et~al.(2017)Oates, Kasza, Simpson, and Forbes]{oates2017}
Oates, C., Kasza, J., Simpson, J., and Forbes, A.
\newblock Repair of partly misspecified causal diagrams.
\newblock \emph{Epidemiology}, 28, 2017.

\bibitem[OpenAI(2023{\natexlab{a}})]{openai2023gpt4}
OpenAI.
\newblock Gpt-4 technical report, 2023{\natexlab{a}}.

\bibitem[OpenAI(2023{\natexlab{b}})]{openai2023gpt}
OpenAI, R.
\newblock Gpt-4 technical report.
\newblock \emph{arXiv}, pp.\  2303--08774, 2023{\natexlab{b}}.

\bibitem[Ouyang et~al.(2022)Ouyang, Wu, Jiang, Almeida, Wainwright, Mishkin,
  Zhang, Agarwal, Slama, Ray, et~al.]{ouyang2022training}
Ouyang, L., Wu, J., Jiang, X., Almeida, D., Wainwright, C., Mishkin, P., Zhang,
  C., Agarwal, S., Slama, K., Ray, A., et~al.
\newblock Training language models to follow instructions with human feedback.
\newblock \emph{Advances in Neural Information Processing Systems},
  35:\penalty0 27730--27744, 2022.

\bibitem[Peters et~al.(2017)Peters, Janzing, and
  Sch{\"o}lkopf]{peters2017elements}
Peters, J., Janzing, D., and Sch{\"o}lkopf, B.
\newblock \emph{Elements of causal inference: foundations and learning
  algorithms}.
\newblock The MIT Press, 2017.

\bibitem[Runge et~al.(2019)Runge, Bathiany, Bollt, Camps-Valls, Coumou, Deyle,
  Glymour, Kretschmer, Mahecha, Mu{\~n}oz-Mar{\'\i},
  et~al.]{runge2019inferring}
Runge, J., Bathiany, S., Bollt, E., Camps-Valls, G., Coumou, D., Deyle, E.,
  Glymour, C., Kretschmer, M., Mahecha, M.~D., Mu{\~n}oz-Mar{\'\i}, J., et~al.
\newblock Inferring causation from time series in earth system sciences.
\newblock \emph{Nature communications}, 10\penalty0 (1):\penalty0 2553, 2019.

\bibitem[Sachs et~al.(2005)Sachs, Perez, Pe'er, Lauffenburger, and
  Nolan]{sachs2005causal}
Sachs, K., Perez, O., Pe'er, D., Lauffenburger, D.~A., and Nolan, G.~P.
\newblock Causal protein-signaling networks derived from multiparameter
  single-cell data.
\newblock \emph{Science}, 308\penalty0 (5721):\penalty0 523--529, 2005.

\bibitem[Scheines et~al.(1998)Scheines, Spirtes, Glymour, Meek, and
  Richardson]{scheines1998tetrad}
Scheines, R., Spirtes, P., Glymour, C., Meek, C., and Richardson, T.
\newblock The tetrad project: Constraint based aids to causal model
  specification.
\newblock \emph{Multivariate Behavioral Research}, 33\penalty0 (1):\penalty0
  65--117, 1998.

\bibitem[Scutari(2010)]{bnlearn}
Scutari, M.
\newblock Learning bayesian networks with the {bnlearn} {R} package.
\newblock \emph{Journal of Statistical Software}, 35\penalty0 (3):\penalty0
  1--22, 2010.
\newblock \doi{10.18637/jss.v035.i03}.

\bibitem[Spiegelhalter \& Cowell(1992)Spiegelhalter and Cowell]{child}
Spiegelhalter, D.~J. and Cowell, R.~G.
\newblock Learning in probabilistic expert systems.
\newblock pp.\  447--466, 1992.

\bibitem[Spirtes et~al.(2000)Spirtes, Glymour, and
  Scheines]{spirtes2000constructing}
Spirtes, P., Glymour, C., and Scheines, R.
\newblock Constructing bayesian network models of gene expression networks from
  microarray data.
\newblock 2000.

\bibitem[Spirtes et~al.(2013)Spirtes, Meek, and Richardson]{FCI}
Spirtes, P.~L., Meek, C., and Richardson, T.~S.
\newblock Causal inference in the presence of latent variables and selection
  bias.
\newblock \emph{arXiv preprint arXiv:1302.4983}, 2013.

\bibitem[Tu et~al.(2023)Tu, Ma, and Zhang]{tu2023causaldiscovery}
Tu, R., Ma, C., and Zhang, C.
\newblock Causal-discovery performance of chatgpt in the context of neuropathic
  pain diagnosis.
\newblock 2023.

\bibitem[Willig et~al.(2022)Willig, Ze{\v{c}}evi{\'c}, Dhami, and
  Kersting]{willig2022can}
Willig, M., Ze{\v{c}}evi{\'c}, M., Dhami, D.~S., and Kersting, K.
\newblock Can foundation models talk causality?
\newblock \emph{arXiv preprint arXiv:2206.10591}, 2022.

\bibitem[Zheng et~al.(2018)Zheng, Aragam, Ravikumar, and Xing]{zheng2018dags}
Zheng, X., Aragam, B., Ravikumar, P.~K., and Xing, E.~P.
\newblock Dags with no tears: Continuous optimization for structure learning.
\newblock \emph{Advances in neural information processing systems}, 31, 2018.

\end{thebibliography}
\bibliographystyle{icml2022}

\newpage
\appendix
\onecolumn

\section{Additional Experimental Results}\label{app:more-results}


\begin{figure}[h!]
    \centering
    \includegraphics[width=\textwidth, clip, trim=0 0 0 40]{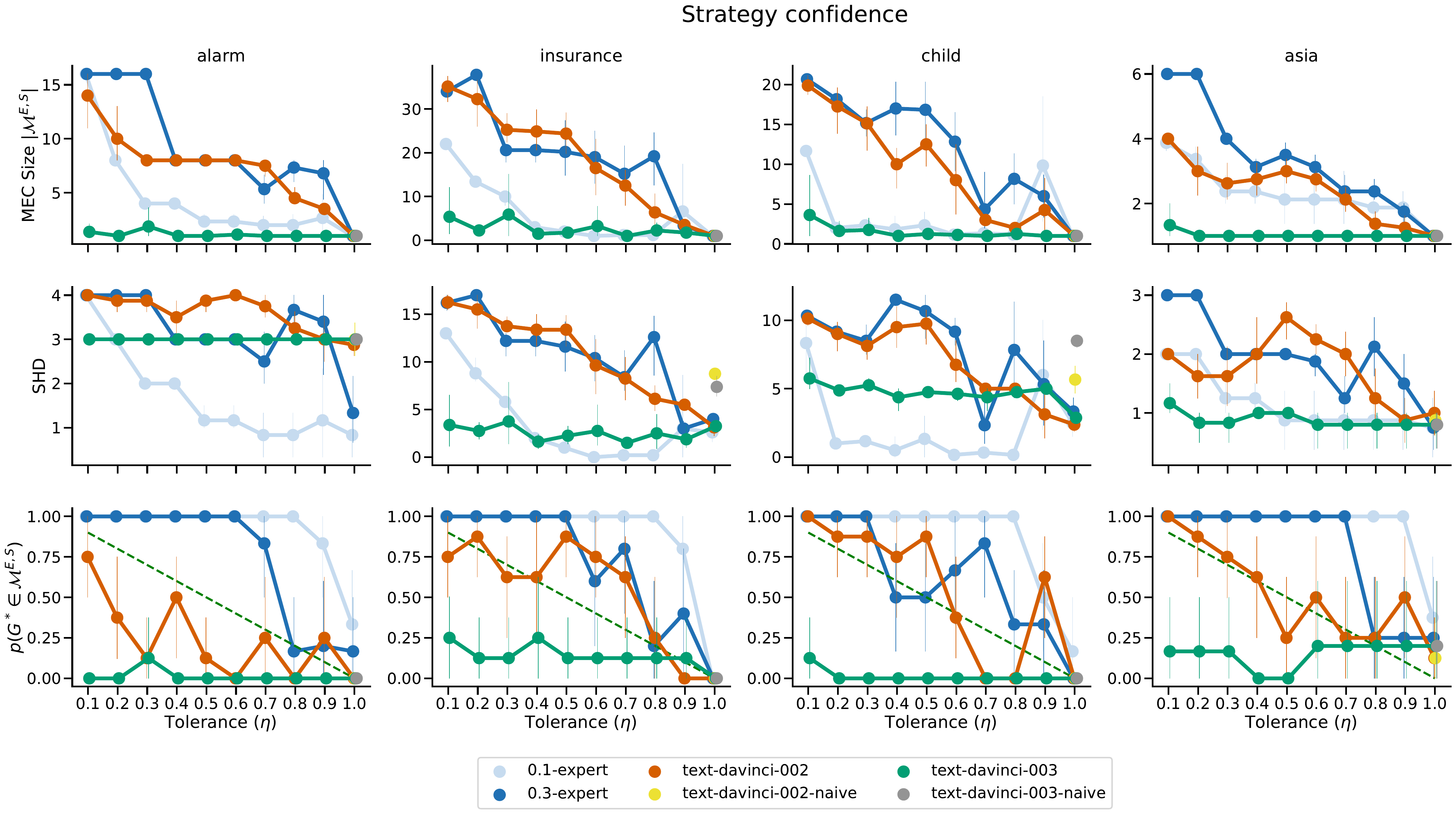}
    \caption{All results for strategy $S_\text{risk}$. We observe that the MEC size consistently decreases as the tolerance level is increased.} 
    \label{fig:greedy_conf_noisy_expert}
\end{figure}

\begin{figure}[h!]
    \centering
    \includegraphics[width=\textwidth, clip, trim=0 0 0 40]{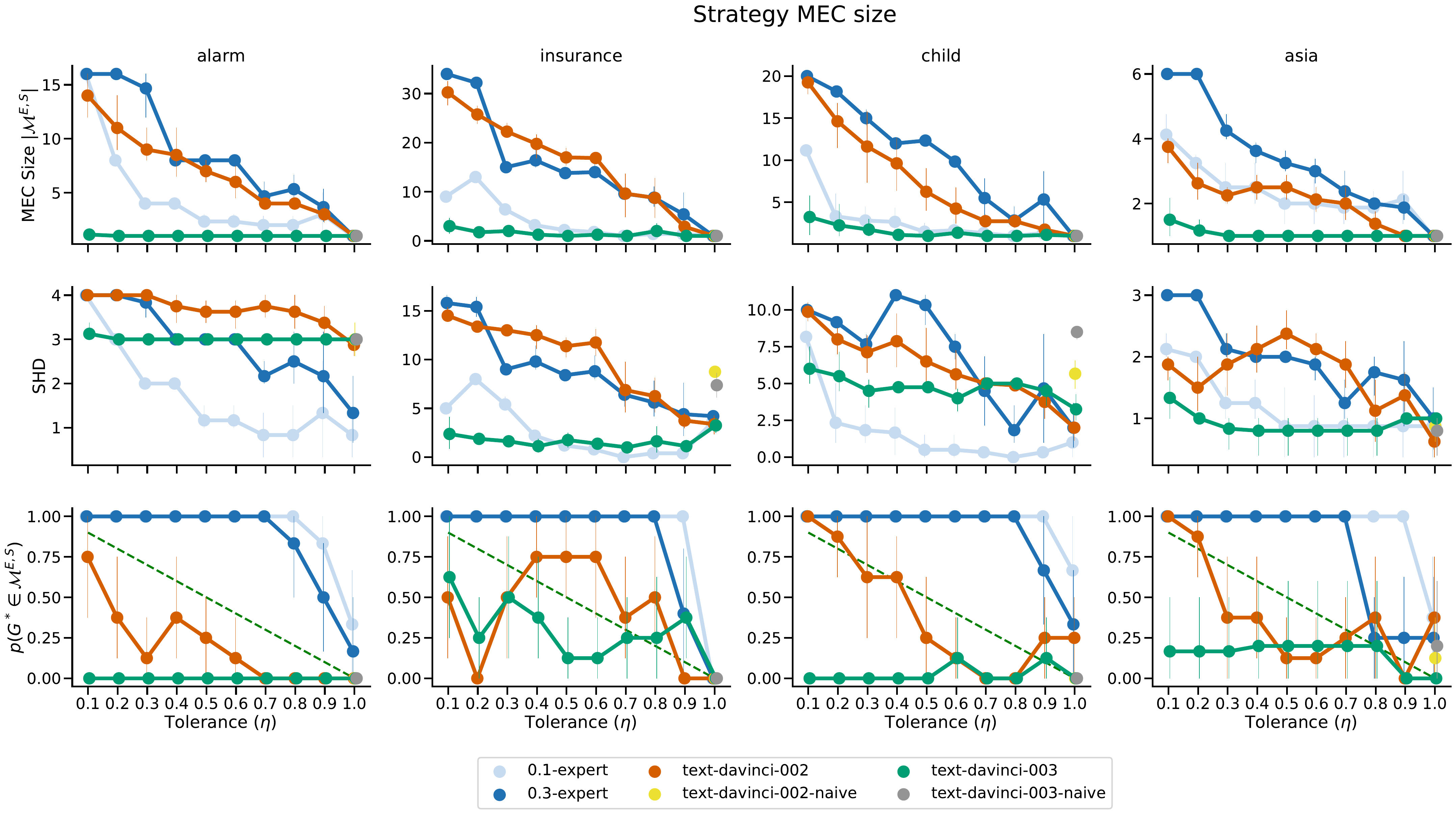}
    \caption{All results for strategy $S_\text{size}$. We observe that the MEC size consistently decreases as the tolerance level is increased.} 
    \label{fig:greedy_mec_noisy_expert}
\end{figure}


\begin{table}[h]
\centering
\caption{Characteristics of included causal networks}
\label{tab:dataset-info}
\begin{tabular}{cccc}
\hline
\textbf{Dataset} & \textbf{\# Nodes} & \textbf{\# Edges} & \textbf{Parameters}\\ \hline
Asia & 8 & 8 & 18\\
CHILD & 20 & 25 & 230\\
Insurance & 27 & 52 & 1008\\
ALARM & 37 & 46 & 509\\ \hline
\end{tabular}
\end{table}

\section{Implementation details}

The code for this work is available at \href{https://github.com/StephLong614/Causal-disco}{https://github.com/StephLong614/Causal-disco}.

\section{Details for the BnLearn causal Bayesian networks}\label{app:bnlearn}

\textbf{Acquisition:~} All Bayesian network structures were acquired from \href{https://www.bnlearn.com/bnrepository/}{https://www.bnlearn.com/bnrepository/}.

\textbf{Variable metadata:~} For each of the included causal Bayesian networks, we extracted a \emph{code book}, i.e., a list of variable names and an associated description (e.g., 'birth asphyxia': 'lack of oxygen to the blood during birth'), from the associated original paper. For instance, for the \texttt{CHILD} network, this information was extracted from \citet{child}.
All code books are available at: \href{https://github.com/StephLong614/Causal-disco/tree/main/codebooks}{https://github.com/StephLong614/Causal-disco/tree/main/codebooks}.

\begin{figure}[h!]
    \centering
    \includegraphics[width=0.5\textwidth]{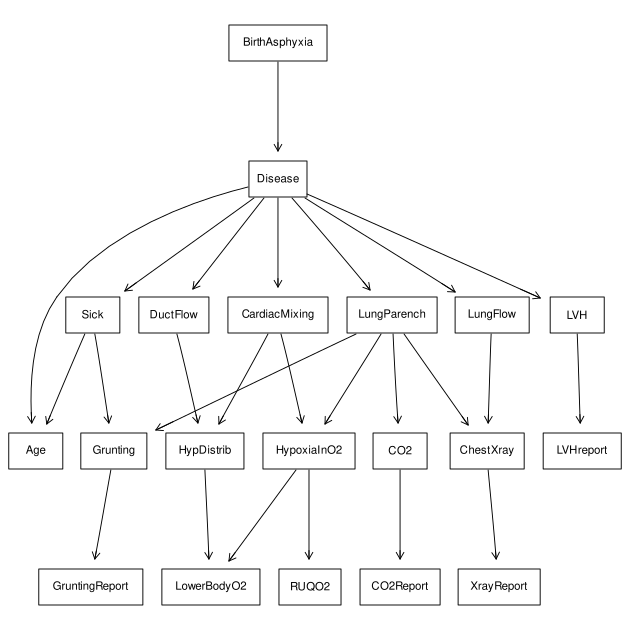}
    \caption{\textbf{CHILD} Bayesian network which represents the presentation of six possible conditions that lead to ``blue babies'' i.e., birth asphyxia \cite{child}. Abbreviations: \textit{LungParench = Lung parenchyma, LVH = left ventricular hypertrophy; HypDistrib = hypoxia distribution; RUQO2 = right upper quad oxygen level.}}
    \label{fig:CHILD}
\end{figure}

\textbf{Metadata pitfalls:~~} Certain Bayesian networks contain edge orientations between variable pairs that appear incongruent with intuitive reasoning. For example, in the CHILD Network (Figure \ref{fig:CHILD} ), the edge orientation between \textit{disease} and \textit{age} exhibits a counterintuitive direction: $disease \rightarrow age$. Implying the causal relationship of ``disease causes age'' rather than the more intuitive and expected ``age causes disease''.

\section{Details for LLM-based experts}\label{app:llm-experts}

\subsection{Querying for edge orientations}

In order to obtain a probability distribution over the orientations of an edge, we use a prompt similar to \citet{bai2022constitutional}. We use the following prompt format:

\texttt{
Among these two options which one is the most likely true:\\
(A) \{$\mu_i$\}\ \{$\text{verb}_k$\} \{$\mu_j$\} \\
(B) \{$\mu_j$\}\ \{$\text{verb}_k$\} \{$\mu_i$\} \\
The answer is: 
}

where $\text{verb}_k$ is randomized at each decision and the variables .

For example, if we wanted to elicit a prediction for the direction of an edge between variables
with metadata $\mu_i$: ``lung cancer'', $\mu_j$: ``cigarette smoking'', and causation verb $\text{verb}_k$: ``causes'' we would use the following prompt:

\texttt{
Among these two options which one is the most likely true:\\
(A) lung cancer causes cigarette smoking\\
(B) cigarette smoking causes lung cancer' \\
The answer is: 
}

We then compute the log probability of the responses \texttt{(A)} and \texttt{(B)}, and use the softmax to obtain a probability distribution over the directions of the edge~\citep{kadavath2022language}. Since we rely on scoring, instead of generation, the output of the LLM-expert is deterministic given a fixed prompt. 
To foster randomness in the LLM-expert outputs, we randomly draw $\text{verb}_k$ from the following verbs of causation:
provokes, triggers, causes, leads to, induces, results in, brings about, yields, generates, initiates, produces, stimulates, instigates, fosters, engenders, promotes, catalyzes, gives rise to, spurs, and sparks.

\section{Causal Bayesian networks included}
We included the following causal Bayesian networks in this work. All were extracted from the \emph{bnlearn} Repository \href{https://www.bnlearn.com/bnrepository/}{https://www.bnlearn.com/bnrepository/}. 

\begin{figure}[h!]
    \centering
    \includegraphics[width=0.5\textwidth]{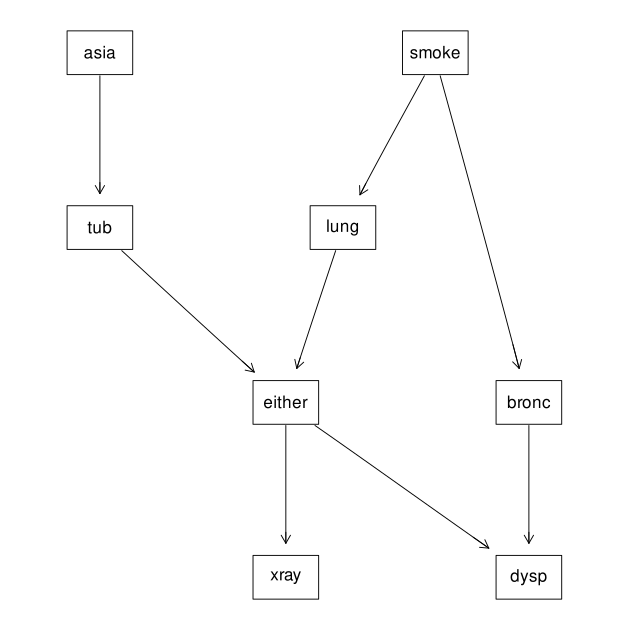}
    \caption{\textbf{Asia} Bayesian network representing a fictitious medical illustrating possible causes of shortness-of-breath (dyspnoae) \cite{asia}. Abbreviations: \textit{asia = visit to Asia?; tub = Tuberculosis; either = either tuberculosis or lung cancer; lung = lung cancer; bronc = bronchitis; dysp = dyspnoae.}
    ]}
    \label{fig:asia}
\end{figure}

\begin{figure}[h!]
    \centering
    \includegraphics[width=0.5\textwidth]{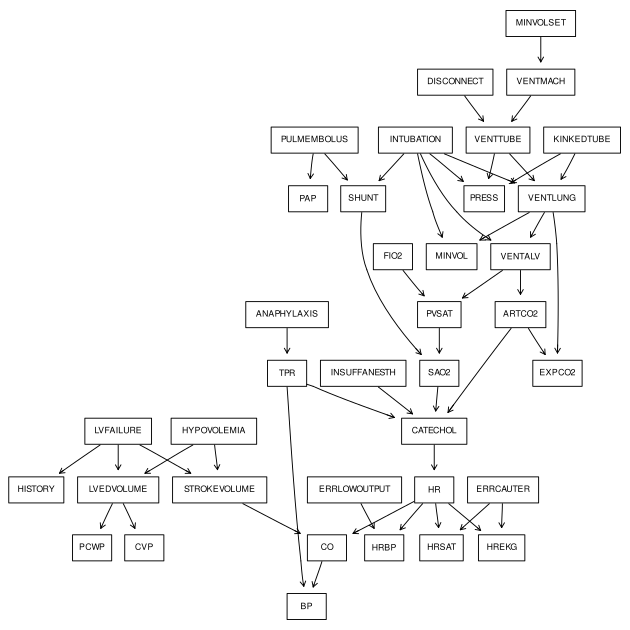}
    \caption{\textbf{ALARM} Bayesian network representing a diagnostic application for patient monitoring which includes 8 diagnoses, 16 findings, and 13 intermediate variables \cite{alarm}. Abbreviations: \textit{MINVOLSET = minute ventilation; VENTMACH = ventilation machine; PULMEMBOLOUS = pulmonary embolism; PAP = pulmonary artery pressure; FIO2 = fraction of inspired oxygen; MINVOL = minute volume; VENTALV = alveolar ventilation; PVSAT = pulmonary artery oxygen saturation ; ARTCO2 = arterial CO2; TPR = total peripheral resistance; SAO2 = oxygen saturation; EXPCO2 = expelled CO2; LVFAILURE = left ventricular failure; CATECHOL = catecholamine; LVEDVOLUME = left ventricular end-diastolic volume; HR = heart rate; ERR = error; PCWP = pulmonary capillary wedge pressure; CVP = central venous pressure; CO = cardiac output; HRBP = rate blood pressure}; HRSAT = heart rate saturation}
    \label{fig:alarm}
\end{figure}

\begin{figure}[h!]
    \centering
    \includegraphics[width=0.5\textwidth]{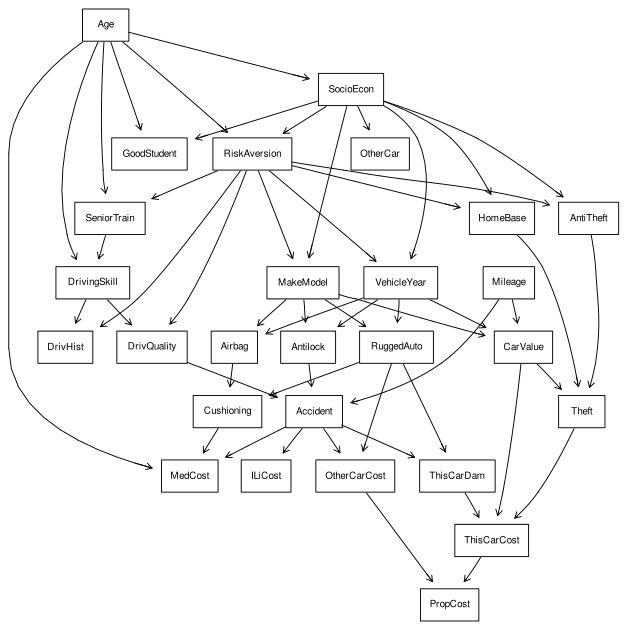}
    \caption{\textbf{Insurance} Bayesian network illustrating factors that affect expected claim costs for a car insurance policyholder \cite{insurance}. Abbreviations: \textit{DrivHist = driving history; ILiCost = insurance liability cost; PropCost = property cost}}
    \label{fig:insurance}
\end{figure}

\end{document}